# Is calibration a fairness requirement?

An argument from the point of view of moral philosophy and decision theory

Michele Loi

Politecnico di Milano, Dipartimento di Matematica, michele.loi@polimi.it

Christoph Heitz

Zurich University of Applied Sciences, christoph.heitz@zhaw.ch

In this paper, we provide a moral analysis of two criteria of statistical fairness debated in the machine learning literature: 1) calibration between groups and 2) equality of false positive and false negative rates between groups. In our paper, we focus on *moral* arguments in support of either measure. The conflict between group calibration vs. false positive and false negative rate equality is one of the core issues in the debate about group fairness definitions among practitioners. For any thorough moral analysis, the meaning of the term "fairness" has to be made explicit and defined properly. For our paper, we equate fairness with (non-)discrimination, which is a legitimate understanding in the discussion about group fairness. More specifically, we equate it with "prima facie wrongful discrimination" in the sense this is used in Prof. Lippert-Rasmussen's treatment of this definition. In this paper, we argue that a violation of group calibration may be unfair in some cases, but not unfair in others. This is in line with claims already advanced in the literature, that algorithmic fairness should be defined in a way that is sensitive to context. The most important practical implication is that arguments based on examples in which fairness requires between-group calibration, or equality in the false-positive/false-negative rates, do no generalize. For it may be that group calibration is a fairness requirement in one case, but not in another.

## 1 Introduction

In this paper, we provide a moral analysis of two criteria of statistical fairness debated in the machine learning literature: 1) calibration between groups and 2) equality of false positive and false negative rates between groups. The measures considered in this paper appear under different designations in different sources: in [3] these measures are called, respectively, *sufficiency* and *separation*. In [5] they are called "conditional use accuracy equality" and "conditional procedure accuracy equality". In [12], the first is called "equalized odds" and in [8] the second is called "well-calibration". In the following, we refer to the measures with the expressions "equality of false positive and false negative rates" and "between-group calibration".

In our paper, we focus on *moral* arguments in support of either measure. The conflict between group calibration vs. false positive and false negative rate equality is one of the core issues in the debate about group fairness definitions among practitioners. Take for example the debate on the COMPAS tool. The algorithm used to decide whether to release inmates on parole has been judged, by some, to be unfairly biased against minority groups [2], based on inequality in false positive and false negative rates between the two groups. However, it has shown that the COMPAS tool achieves between-group calibration to a reasonable degree [6], thus fulfilling an established fairness criterion. The text in [2] provides no clear argument why unequal false positive and false negative rates should be viewed as discriminatory.

For any thorough moral analysis, the meaning of the term "fairness" has to made explicit and defined properly. For our paper, we equate fairness with (the absence of) discrimination, which is a legitimate understanding in the discussion about group fairness. More specifically, we equate it with "prima facie wrongful discrimination" (see 2.4). We base our paper on an influential definition of discrimination from the philosophical literature which has been defended at length in [19], and we use this as a conceptual foundation of deciding whether or not discrimination happens.

The paper is structured as follows: 2 provides the definitions of the considered statistical fairness definitions this paper is about, and the used definition of discrimination [19] is reviewed. In 3, we review related work. In 4.2 and 4.3. we sketch the general moral analysis of the two cases, the first (4.2) concerning a case in which the violation of between-group calibration implies unfairness and the second (4.3.) concerning a case in which it does not. Both arguments rely on a premise about the way the utility of individuals of two groups depend on the qualities of prediction scores. These premises are rather technical and rely on the analysis and additional assumptions provided in section 4 and 5, respectively.

## 2 Definitions

### 2.1 Definition of between-group calibration

In this paper, by "between-group calibration" we mean the following condition between variables (called sufficiency in [3]):

$P(Y = y \mid A = a, R = r) = P(Y = y \mid R = r)$,
for all values $r, a$, and $y$

Y stands for an outcome of interest that is unknown at the point of decision making, e.g., whether a parolee will reoffend. For simplicity of exposition, we consider a Y that is binary (e.g., y=1 when the parolee does reoffend, y=0 otherwise), but our arguments in this paper can easily extended to non-binary cases. R is a prediction of Y which is used for making or informing a decision. R may have only a few discrete values (e.g., 1 and 0 in a binary classifier) or values in a continuum (e.g., any value between 1 and 0). The former case represents a point prediction, where the latter represents a probabilistic prediction, typically expressed by a score. Later in the paper, we will use *p* instead of *r* for cases where we the predictor is to be understood as a probability. The variable A denotes the group to which the person about the prediction is made belongs, e.g., a = male. Typically we are interested in "socially salient groups" [18] such as sex, gender, ethnicity, religion, possibly age, etc... not groups deprived of social meaning such as "people with an even house number". We assume for argument's sake that some account exists of what makes a group socially salient in the required sense, but we offer no substantive moral view for the problem of its grounding.

We use the standard notation of capital letters (e.g. "Y") for the random variable itself, and small letters (e.g. "y") for a realized value of this variable. What the equality sign says is that the probability that the variable Y takes a certain value is the same for all groups A, given a specific value of r of R, for all possible r.

Notice that this definition of between-group calibration is different to the classical ML definition of calibration used for probabilistic predictors, meaning that $P(Y = 1 \mid R = r) = r$. We only require that the probability of Y=1 and Y=0 is equal for both groups, given a specific value *r*. Notice that, in general, calibration depends on the considered population: a score may be calibrated for one group but not for a different group, and a score may be calibrated for the whole population but not for subgroups. For example, the score R described as "probability that a person released from prison reoffends" may be calibrated for the whole population, but not for men. For the example of r=0.8, we may find that from 100 people with this score, 80 of them reoffended. But if we look at "the probability that a *man* released from prison reoffends", the same score may violate calibration, e.g., if out of the same 100 people 81 were males and all but one of them reoffended. Thus, calibration can be relative to a group, and the difference to between-group calibration is in the non-comparative nature of its definition. One should distinguish between-group calibration (the same probability for different groups) from the score being calibrated (the probability corresponds to the frequency) for all groups seperately. Following Kleinberg and colleagues [17], we refer to the latter as "calibration within groups".

## 2.2 Definition of Equality of false positive and false negative rate

Let us introduce the condition known as *separation* in [3]. Separation is achieved if

Separation means that, if we condition on a specific value of Y (e.g. Y=1), then the probability of having a predicted value of r is equal for each group. For example, if we take all prisoners who will eventually reoffend, the probability of a positive prediction (R=1) is equal for blacks (A=1) and whites (A=0).

Typically separation is considered when the score *R* is a binary score, $r \in \{0,1\}$ which can be interpreted as a positive or negative prediction of the event. In such cases, R is often denoted by the symbol $\hat{Y}$, expressing that it is a point prediction of the unknown Y. Thus, R = $\hat{Y}$ = 1 indicates for example the prediction that the parolee will reoffend and R = $\hat{Y}$ = 0 the prediction that the parolee will not reoffend. Separation implies then equality of false positive and false negative rates of the binary classifier, which can be expressed as equality in the proportions (b/(b+d)) and (c/(a+c)) in the confusion matrixes below describing the performance of the classifier in different demographic groups *a*.

**Table 1. A confusion matrix for a recidivism classifier**

|  |  | Outcomes (Y) | |
|---|---|---|---|
|  |  | Reoffending (Y=1) | Non reoffending (Y=0) |
| Predictions $\hat{Y}$ | Will reoffend $\hat{Y}$ = 1 | (a)True Positives | (b)False Positives |
|  | Will not reoffend $\hat{Y}$ = 0 | (c) False Negatives | (d) True Negatives |

If we suppose that a positive/negative prediction automatically converts to a given decision (e.g., to deny or grant parole) and we indicate the decision with the variable D, we can rewrite the confusion matrix of Table 1 as the decision matrix below.

**Table 2. A decision matrix for a recidivism classifier**

|  |  | Outcomes (Y) | |
|---|---|---|---|
|  |  | Reoffending (Y=1) | Non reoffending (Y=0) |
| Decision $D$ | Deny parole D = 1 | (a)True Positives | (b)False Positives |
|  | Award parole D = 0 | (c) False Negatives | (d) True Negatives |

Even though true positives, etc…, are normally used to refer to the proportion of correct and incorrect predictions, we shall sometimes use the same terms to refer to the proportion of the corresponding decisions, given the assumption above. Whenever we do so we shall write it explicitly.

## 2.3 Impossibility theorems

Imperfectly accurate predictors cannot satisfy between-group calibration and equality of false positives and false negatives when the base rate distribution of Y (that is, the proportion (a+c) / (a+b+c+d) in table 1) is unequal between the groups [3,17]. Analogous impossibilities follow from this fundamental result (see also [13]): it is impossible to equalize the false positive / false negative rates when scores *p* are required to be between-group calibrated [8] and when classifiers are required to satisfy conditional procedure accuracy equality [5].

## 2.4 Definition of (un)fairness

In this paper, by unfairness we mean *the existence of prima facie wrongful* discrimination, a concept that we will analyze below [19], 2.5. This is a stipulation of our paper, as it gives the term fairness a well-defined meaning. Those who disagree may simply translate all the claims made in the abstract and introduction about fairness as claims that only concern the absence of prima facie wrongful discrimination.

## 2.5 Definition of prima facie wrongful discrimination

We take our definition of discrimination from the philosophical literature [19]. This definition is designed to select a concept of discrimination, more precisely labelled as "group discrimination", for which it is generally reasonable to think that, if something counts as discrimination in this sense, it may probably be wrong in the way that is specific to all discrimination cases that are wrong because they are instances of discrimination. [19] refers to this as that concept of discrimination that is "prima-facie" wrong. The author concedes that there may be other uses of the word "discrimination", such that they do not denote something which is generally wrong, or that, when it is, it is wrong *because* it is discrimination. In this definition, Φ-ing is the act that may be discriminatory and, because it is discriminatory, is prima facie morally wrong. X is the decision maker who may be said to engage in discrimination. Y is the person discriminated against. Z is the person whose treatment provides the standard against which the treatment of Y is judged to be discriminatory. A property P distinguishes Y from Z.

> Prima facie wrongful discrimination (DEF)
>
> X discriminates against Y in relation to Z by Φ-ing if, and only if, (i) there is a property, P, such that (X believes that) Y has P and (X believes that) Z does not have P, (ii) X treats Y worse than Z by Φ-ing, (iii) it is because (X believes that) Y has P and (X believes that) Z does not have P that X discriminates against Y in relation to Z by Φ-ing (iv) P is the property of being member of a certain socially salient group (to which Z does not belong), and (v) Φ-ing is a relevant type of act etc., and there are many acts etc. of this type, and this fact makes people with P (or some subgroup of these people) worse off relative to others, or Φ-ing is a relevant type of act etc., and many acts etc. of this type would make people with P worse off relative to others, or X's Φ-ing is motivated by animosity towards individuals with P or by the belief that individuals who have P are inferior or ought not to intermingle with others. [19:1.8]

We regard this definition as one of the most influential state-of-art definitions of discrimination [20].[1] We do not review alternative concepts of discrimination and theories of the wrongfulness of discrimination [1,e.g., 15]. The steps necessary to connect in a rigorous way statistical fairness definitions and philosophical ones are already so complex that a paper-length treatment of the topic has to proceed one concept at the time.

While it is hard to do justice to the complexity of the moral reasoning behind each single condition in Lippert-Rasmussen's definition proposal, let us try to clarify his core points.

First, consider (iii). This requires that discrimination happens for a specific reason, namely, the fact that the discriminated person has P, or is believed to have P (neither possibility is excluded by definition). For example, a non-black person can be discriminated by someone who believes him to be black, and a black person can be discriminated indirectly by a rule that unjustifiably makes it harder for black people to obtain an advantage. Clauses (i) and (ii) tell us more about what P is: as (v) entails that discrimination requires us to assess for unequal treatment of different groups (more details to follow), clause (i) tells us that it is precisely this P that determines how we differentiate the groups to compare. In the typical cases discussed in the machine learning literature, P is membership to a group, which we have indicated with A in sections 2.1 and 2.2. Also in Lippert-Rasmussen's definition of discrimination, the property P must be the membership to a socially salient group, by clause (iv''). Clause (ii) explicitly requires that a discrimination claim rests on a comparison with respect to the *treatment* that individuals receive, with one treatment qualified as *worse* than the other. Let us now focus on (v). This condition is the one that plays the most important role in relating statistical measures to discrimination.

---

[1] In the light of the fact that some alternative accounts of the moral wrongness of discrimination [10, 15] have elements that can be mapped into these. In particular, they also imply a comparative element and disadvantageous treatment (which may not coincide with harm). The precise way in which those elements are expressed may differ. For example, the account of discrimination in [10] requires treating the discriminated person worse (or better) than others in some respect *on the basis of* some trait possessed or thought to be possessed by the first person, where "on the basis of" plays a similar function as "because" in (iii), but is not quite the same relation.

The reasoning for (v) is the most complex. The other conditions have the counter-intuitive consequence of implying that heterosexual people engage in discrimination merely by virtue of being heterosexual, i.e., by virtue of avoiding sex with people (of the same sex) by whom they are *not* sexually attracted and with whom they are incapable of falling in love. (Remember that the goal of the definition is to nail down a concept of discrimination such that if something is discrimination in that sense it can be and is generally wrong for that reason.) Clause (v) implies that heterosexuals engage in (prima facie morally wrongful) discrimination if they believe that homosexuals are inferiors and should not intermingle with others, or if they are animated by animosity against them, but not otherwise. For the first disjunct of (v) is typically not satisfied: as observed in [19] "homosexuals would not be better off relative to others if heterosexual men were neither to reject advances from people in whom they are not sexually interested, nor more likely to fall in love with people of the opposite sex" [19:1.5].

## 3 Related work

Here, we review *exclusively* the debate on between group calibration and FP-FN rate equality in which a moral argument (or something sufficiently close to it) has been given explicitly. These two mathematical constraints are arguably some of the most heavily discussed and most heavily used metrics in discussions of the bias of algorithms. Yet, moral arguments explaining in which sense they measure fairness in a morally relevant sense and are rightly called "fairness requirements" have been rarely given [7, 22, 13, 11]. Some arguments appealing to legal and economic notions of discrimination appear however to be highly relevant to these moral questions.

For example [16] has argued that "equal predictive accuracy also relates primarily to questions of belief and not to questions of action". The first part of this claim is in line with the view, supported also by [9], that calibration is fundamentally valuable because it is required by the equality of epistemic or semantic value of a statistical indicator: "Calibration ensures that risk scores s(x) mean the same thing for all protected groups—for example, that white and black defendants given a risk score of 7 indeed recidivate at comparable rates." Notice that [9] also claim that using decision rules with the same risk thresholds for different groups satisfies legal and economic definitions of non-discrimination. For example, if scores are calibrated, denying parole to all individuals with a reoffending risk higher than a given threshold equalizes the expected benefit from every parole decision, on the assumption that the expected harm of releasing a future reoffender is the same for all future reoffenders [9]. Given the assumption, this equalization is required to avoid taste-based discrimination by the decision-maker in Becker's sense [4]. This view contradicts the one defended by [16], to the extent that achieving equality of false-positive and false-negative rates with calibrated scores requires using different thresholds for different groups [12]. [16] argues for the equality of false-positive/negative rates when their violation implies that the moral importance of false-positive vis-à-vis false-negatives, e.g., of wrongly punishing vs. wrongly avoiding punishment, depends from group membership. By advocating for this, [16] rejects calibration as fairness requirement. By contrast, [16], [23, 13] show, with a very persuasive mental experiment, that equality in the false-positive/false-negative rates cannot be plausibly considered a *necessary condition* of fairness (in all logically possible situations). The two views are compatible, since [16] does not argue for the strong position concerning a necessary condition, that the argument by [13, 23] shows to be implausible.

The view we defend here differs from the above alternatives, because it is based on a different, highly influential account of discrimination from analytic philosophy. It is highly relevant for anyone concerned with fairness in machine learning (Fair ML) to note that, starting from this established account one can show that a group fairness definition (calibration) which is morally relevant in one context may not be relevant in a different one. This idea may be regarded to be uncontroversial or even commonsense in the Fair ML literature. Yet, there are few analyses showing that (and how) the application of one general conception of fairness may lead to different group fairness definitions, depending on the context.[2] Our paper shows, in contrast to [16], that between-group calibration can clearly be relevant to questions of action and its violation may lead to group-discrimination and unfairness between groups. In contrast to [9], our argument concerning the use of calibrated scores does not relate to the "tastes" (in the sense of [4], i.e., discriminatory preferences) of the decision maker, but explains how individuals can be *negatively affected by* non-calibrated scores.[3] Compatibly with views defended in [16] it also shows that it is not *necessary* that a violation of calibration will produce group discrimination.

## 4 The argument

### 4.1 A very short sketch

First, we provide a recommendation system as an example of prediction-based decision making where scores violate between-groups calibration and argue that, as a result, the agent providing the non-calibrated scores discriminates against one group. This is because, as we shall argue, the fact that the scores violate between-group calibration makes people of one group worse off in expectations than people of a suitable comparison group. Then, we provide an example where a decision maker uses scores that are not between-group calibrated and argue that in this case no group is discriminated against. In the second case, what makes it the case that there is no discrimination is the fact that no group is worse off as a result of using scores lacking calibration. Incidentally, this is because equality in the false positive and false negative rates is achieved, and this, *in this specific case*, implies that no group is made worse off by the use of the scores.

---

[2] Two such examples are [14, 22].
[3] Admittedly, according to Becker's [20] taste-based analysis, discrimination produces harmful effects on the members of some group. But here we discuss cases, e.g., the recommender system case and the judge case, which are, in our view, not easily reconciled with Becker's concept of taste-based discrimination.

In this way we achieve what we promised in the introduction, namely, an example in which the violation of between-group calibration creates an instance of unfairness and an example in which it does not. This proves that there are situations where between-group calibration indeed is necessary for avoiding discrimination, which justifies to call between-group calibration a valid fairness measure. But there are also cases in which between-calibration can be violated without implying that discrimination occurs. To the extent that the absence of *prima facie* wrongful discrimination, as defined here, is what "fairness" consists in, we can show how the relevance of fairness constraints depend on the specific way predictive scores are used.

## 4.2 The first argument: violating calibration can be unfair

In the first example, the violation of group calibration amounts to discrimination against members of one group. In this example, decision-makers of two groups, men and women, obtain a score from a trusted advisor that they use to make decisions, where these decisions are exclusively for their own benefit. The score is calibrated *within* the group of women. (For the group of women, if the score *p* is 0.8 it corresponds to 80% of successes.) The score is not calibrated *within* the group of men, if the score *p* is 0.8 it correspond to some value *other than* 80%. Since the same score *p* corresponds to different probabilities P[Y=1] in the case of men and women, *between-group* calibration is also violated.

The example we imagine is as follows: Ann and Bob want to watch a movie and rely on a recommendation system for this purpose. The practical question each of them faces concerns whether to spend time (a scarce resource with an opportunity cost) watching a given movie or not. They both interpret the score, p, attached to a movie recommendation, as an indication of how likely it is that they will enjoy it.

We will identify a rule of rational decision making which both Ann and Bob should follow in order to maximize their expected utility, where they are both assumed to believe that the score *p* communicates the probability that they will enjoy the movie. Then we will argue that if the scores Ann receives are within-group calibrated while those Bob receives are not, Bob will be worse off than Ann *in expectation* in the generality of cases. This generalizes to men and women receiving scores when it is generally the case that the scores are within-group calibrated for women but not for men. If scores are within-group calibrated for one group but not for the other, it follows that, necessarily, between-group calibration is violated. Then, the violation of between-group calibration causes discrimination to occur *in this case*.

It is sometimes said that calibration is necessary to guarantee that scores "mean the same thing" when used to describe individuals of the different groups [9, 17]. Thus, our example invents a situation in which people who are users of the scores end up differentially better off when their scores have different meanings. Since the scores are used by men and women to make predictions *about themselves*, in this case, this implies that the violation of group calibration makes people of one group better off than people of the other group.

To be analytically clear about the effects on utility people of one group, from using scores lacking calibration, we identify a decision rule that rational (in the sense of utility-maximizing) individuals are supposed to follow, and then we show that the rule can be used more successfully by decision-makers who rely on calibrated scores. So, if both Ann and Bob adopt this rule, then Ann, being a woman, ends up with higher expected utility than Bob, who is a man, since Ann receives scores that are calibrated, while Bob does not. (It also follows, mathematically, that this is a case in which group calibration is violated.) This arguably amounts to discrimination because it satisfies all the clauses of Lipper-Rasmussen's definition of discrimination. Namely:

- there is a property P, that is being a man, such that Bob has it and Ann does not;
- the recommendation system treats Bob worse than Ann by providing scores to Bob and Ann that are predictive of their enjoyment;
- it is because Bob is a man and Ann is not a man that the recommendation system discriminates against Bob in relation to Ann by providing scores that are gender-sensitive in their predictive quality;
- being a man is the property of being member of a certain socially salient group (to which Ann does not belong);
- providing scores to Bob and Ann is a relevant type of act etc.…, and many acts etc. of this type would make men worse off relative to women.

In section 4 we provide the argument showing that (v) is satisfied. There, we provide a mathematical proof that this rule delivers less utility to Bob (or any other man) than it provides Ann (or any other woman), in expectation. We also show that this may happen also in cases of equality of false-positive and false-negative rates.

Summing up, this argument works as follow. We have designed a situation in which non-discrimination requires between-group calibration of the scores. As the reader will realize by considering the second example, this is far from being a feature of all possible uses of predictive scores.

## 4.3 The second argument: violating calibration is *not* unfair.

Let us now briefly sketch the second argument. Here we deal with a judge who uses predictive scores in which a score R violates group calibration (for men and women), while fulfilling equality in the false positive and false negative rates.[4] We ask whether the use of the scores by the judge is a ground for prima facie wrongful discrimination, as defined. We build this case in such a way that it is plausible that condition (v) of the discrimination definition is not satisfied. To achieve this, we describe a hypothetical case in which the average expected harm caused by the use of the scores in making decisions is equal between two groups, men and women, as a result of the equality in the false positive and false negative rates. Unlike the previous case, here, we assume for simplicity that the user of the score is not affected personally by the quality of his decisions.[5] The only persons affected by decisions taken using predictive scores are the defendants.

---

[4] We assume calibrated scores and group-dependent thresholds. However, as we argue, the coarsened scores (high-risk, low-risk) resulting from different thresholds are not, in turn, calibrated. All we need for this argument is an instance of non-calibrated scores that are not unfair.

Notice, moreover that:
- The score *r*, which is calibrated (for women) or not calibrated (for men) is a coarsened score that obtains from thresholding a score *s*, that is to say, r = 1 if and only if s>T, otherwise r = 0. Such "coarsened" score expresses the probability of the event as high or low, i.e., as a positive or negative prediction.[6]
- We explicitly assume a degree of lack of calibration that is still compatible with the equalization of the false-positive and false-negative rates. Since this is a case in which the unequal benefit/harm for the different groups tracks the false-positive/false-negative rate, it is not a case in which *any* departure from calibration may be compatible with fairness.
- We intentionally select a hypothetical case in which that equality in the false-negative rates implies that the two groups are equally well-off, which, however, is argued to be plausible in this case.
- We also assume we are dealing with an ordinary case in which recidivism prediction is not perfectly accurate and where the baseline of recidivism differs for the different groups, which shows that our result is relevant for the type of situations in which a reason to violate one of the two fairness definitions exists, as explained in 2.3.
- Moreover, we conceive a case in which (realistically, as it often happens) the judge's decision is not motivated by animosity or the belief in the inferiority of one group (this is necessary to satisfy the other disjuncts of condition v in the definition of group-discrimination).

For this case, we argue that the violation of group calibration does *not* amount to discrimination because clause (v) is not fulfilled – assuming, again, that equality in the false-positive and false-negative rates obtain. Since the treatment of individuals amounts to discrimination *only* if all conditions are satisfied, this is sufficient to classify the case as one that is *not* discrimination. In order to show that (v) is *not* satisfied, we have to show that the case of the judge is one in which *none* of the three disjuncts is satisfied. Because of (5), the only factor relevant to assess whether it would be discriminatory to use the scores to make decision is the expected welfare of men and women resulting from Φ-ing, considered as a *type* of act. The act *type* is the (hypothetical, or actual) repetition of the judge's act of using scores violating between-group calibration between men and women (in the specific way described) when making parole decisions. Thus, in order to show that the use of scores (violating between-group calibration) by the judge is not an instance of *prima facie wrongful* discrimination (and therefore not fairness under our definition) it is sufficient to show that (v-1) below is true:

- (v-1) the judge's use of the scores is a relevant type of act etc...., and many acts etc. of this type do not make men worse off relative to women, or conversely.

We interpret (v-1) as equivalent to:

---

[5] Although this may seem a strong assumption, notice that it is actually irrelevant since, even if the judge were harmed by use of non-calibrated scores, *this fact* could not lead to unequal expected harm for women and men.

[6] The coarsened score r (1 or 0) is, however, still a score and, as such, it either satisfies our definition group calibration or not, which then corresponds to predictive value parity for a classifier [9] . It may be objected that focusing on the coarsened score, rather than the raw score itself, is a questionable ad hoc move in this argument. In reply, the choice to focus on the coarsened score is not ad hoc, it is justified because in this case study the well-being of the individuals potentially vulnerable to discrimination is affected by the coarsened score, i.e., by their being classified as future reoffending parolees. It is by virtue of different misclassification rates (expressed by considering coarsened scores) that male or female prison inmates are potentially discriminated by the judge. By contrast, in the first argument, the well-being of the individuals potentially vulnerable to discrimination is affected by the raw score – it was by virtue of the communication of the raw scores that the recommendation algorithm affected the utility of its users, in a way that was harmful to males.

- (v-1-bis) the judge's use of the scores is a relevant type of act etc.…, and many acts etc. of this type cause (or would cause) the same average expected utility for men and women defendants.

# 5 Expected utility in the recommendation case

Ann and Bob receive scores from a movie recommendation system, that are communicated and understood as the probability that the user will enjoy a given movie. We will identify a rule of rational decision making and prove that it maximizes the expected utility of the user only when the scores are calibrated in the non-comparative sense.

Let's assume that the benefit from watching a movie that the viewer likes is U=+1, whereas the benefit from watching a movie that the viewer dislikes is U=-1, because in this case he has wasted time that he could have used for a more pleasurable activity (e.g. watching a better movie, or doing something else such as reading a book). We assume that not watching the movie equals to a benefit of U=0, for both Ann and Bob. We use d = 1 for the decision to watch a movie; d = 0 for the decision not to watch it.

For a given probability p of liking the movie, the expectation value of the viewer's benefit U from watching the movie is

$$E(U) = p \cdot 1 + (1-p) \cdot (-1) = 2p - 1$$

Not watching the movie results in U=0 for all probabilities. So, the optimum decision rule for both Ann and Bob is

d=1, if p>0.5, and d=0 else. This is graphically depicted in Fig. 1.

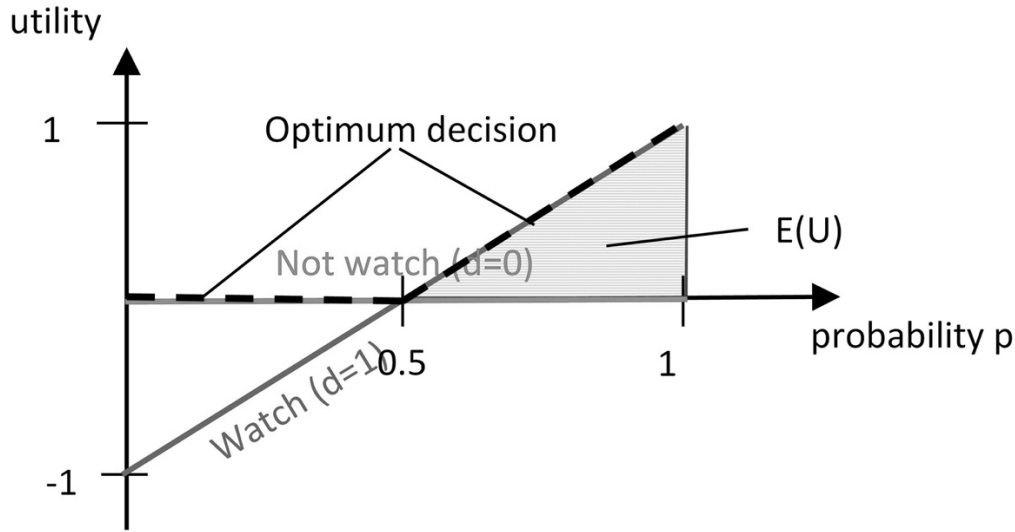

Figure 1: Expected utility as a function of the probability p, for the two options d=0 (horizontal) and d=1(vertical). The optimum decision depends on the probability: It is d=1, if p>0.5, and d=0 else. In the case of a uniformly distributed p, the expected utility under this optimum decision rule is given by the shaded area, denoted by E(U), which is limited by the x-axis (U=0) and the dashed utility function (cmp. Eq. (1).

Now let's assume that Ann and Bob are confronted with a series of movie recommendations, and they always decide according to this decision rule. The average expected utility depends not only on the decision rule, but also on the distribution f(p) of the probabilities p, over the subsequent decisions. For example, if we assume that the probability is equally distributed in the interval [0,1], i.e. f(p)=1, then for the optimum decision rule, the average expected benefit on the long run is:

$$E(U) = \int_{0.5}^{1}(2p-1)f(p)\mathrm{d}p + P(p<0.5] \cdot 0] = 1/4 \quad (1)$$

Or equivalently, the shaded area is equal to the half area of a rectangle with base 0.5 and height 1, that is (0.5*1)/2 = 0.25.

Note that there is no decision rule which yields a larger E(U), and that for realizing this decision rule, the (true) probabilities of each movie has to be known. Note also that the optimum decision rule is the same for all probability distributions f(p), only the achieved long-term benefit differs for different f(p).

Having identified the rule of rational decision making under uncertainty for this case, let's now assume that each recommendation comes with a score s. In Ann cases, the score s always corresponds to the probability p, i.e., s = p. In Bob's cases, this is not always the case, but Bob does not know it. We will now argue that Bob ends up worse off in expectations than Ann.

Let us focus on Ann's case, first. Obviously, Ann is able to generate the maximum utility on the long run, by following the optimum decision rule as described above. This is possible because she has access to the (true) probabilities which are required to take the optimum decision. We now assume that, unlike Ann, Bob does not have access to the true probability p, but instead to a score s. If Bob assumes that the score is equal to the true probability, then his decision rule is d=1, if s>0.5, and d=0 else.

If the score is calibrated, then E(B) =¼, as shown above. We will now analyze the case where the score is not calibrated. For any possible (true) probability p, we have four cases:

1. s<0.5 and p<0.5
2. s>0.5 and p<0.5
3. s<0.5 and p>0.5
4. s>0.5 and p>0.5

In cases 1 and 4, Bob's decision is correct, i.e. he maximizes his expected benefit. In case 2, Bob decides to watch the movie, but receives a negative benefit in expectation, instead of E(U)=0 which would have been the benefit-maximizing decision. In case 3, Bob wrongly decides not to watch the movie, thus creating U=0, where the correct decision would have been to watch the movie, resulting in E(U)>0. This analysis shows that Bob is worse off in the case of a non-calibrated score, unless the non-calibration is such that ALL scores s<0.5 correspond to the cases with

p<0.5, and vice versa. Since this is a highly restrictive requirement, this condition is very unlikely to be met.[7] If we assume that the recommendation algorithm exhibits the same behavior with all women and all men, then not only is Bob treated unequally from Ann by the provision of a non-calibrated score, which satisfies (ii), but men generally are worse off than women, as a result of that type of act, which satisfies (v). Hence, following the above given definition of discrimination, Bob is discriminated *qua* man by this type of violation of between-group calibration.

So far we have discussed a special case of violation of between-group calibration, where we compare a perfectly calibrated score with a non-calibrated score. However, the analysis above also shows that the reason for Bob being worse off is that he takes wrong decisions, which happens for the cases 2 and 3. Loosely speaking, the more such cases happen, the higher is the loss of utility. In other words: the higher the deviation from the reference of a calibrated score, the higher is the loss of utility. This makes clear that two *differently* non-calibrated scores will lead to different expected utilities, unless very specific conditions are met. Thus, most violation of group-calibration, except very rare cases, are discriminatory in expectation.

The loss of utility is generated by wrong decisions. One might assume that the difference of utility loss can be attributed to the difference in the false positive (case 2) or false negative (case 3) rate. However, this is not the case: even if the false positive rates and the false negative rates are equal for both groups, different expected utilities may result. As a simple example, one might imagine a situation where only case 2 appears (only false positives), and the probability of case 2 is equal for both men and women. So, the false positive as well as the false negative rates are equal. However, the false positives for men are concentrated at true probabilities of $p \approx 0.1$, while for women, they are concentrated at $p \approx 0.4$. This means that the number of false decisions is equal for both groups, but the average loss of utility generated by these false decisions is larger for men than for women, which leads to discrimination against men. This is obviously connected to the point in [9,17] that group calibration ensures that scores provide the same degree of evidence when predictions refer to members of different groups.[8]

Summarizing, we have shown that, in this particular setting (a recommendation system, inviting its users of the different relevant groups to make decisions affecting themselves which are based on the non-group-calibrated scores), a violation of between-group calibration leads to discrimination in most cases, unless very specific conditions are met. In addition, we have shown that this discrimination cannot be attributed to a difference of false positive or false negative rates.

---

[7] Another argument: Suppose that the benefit of "not watching the movie" is not 0 but instead another number - e.g. if reading a book as alternative creates a benefit of B=0.5. Then, the optimum decision rule would change, and alternatively the constraint on the non-calibrated score for leading to optimum decisions would change. So, if we consider a (more realistic) case where we do not only have one Bob, but many Bobs using the score for their decision making, with different values of the non-watching alternative, we see that it is very unlikely that an uncalibrated score would not lead to a loss of benefit for some of the Bobs, and it is easy to construct cases where in fact only the calibrated score leads to a maximization of benefit for the whole group of Bobs.
[8] In this case, the score provides a certain (between-group calibrated) degree of evidence when it is used to make predictions about women; but it provides a different (non-between-group calibrated) degree of evidence when used to make prediction about men: that is to say, the score will often either underestimate or overestimate the prospective benefit of watching the movie.

# 6 The judge example

In this section, we provide the argument for:

> (v-1-bis) the judge's use of the scores is a relevant type of act etc...., and many acts etc. of this type cause (or would cause) the same average expected utility for men and women defendants.

We assume that the utility denotes a harm, that is, positive units express *dis*utility, not utility. More specifically, we assume a disutility of value 1 if a defendant who would not re-offend (Y=1) is kept in prison (D=0). This corresponds to the probability P[D=0|Y=1], i.e. the false-negative rate. We assume that in all other cases, no harm is created: Either a defendant is kept in prison correctly, or he is released.[9]

Then, the expectation value of the disutility is

$$E(U) = \begin{cases} P[D=0|Y=1] \cdot 1, & \text{not reoffending (Y=1)} \\ 0, & \text{reoffending (Y=0)} \end{cases}$$

It is clear immediately that, if the false-negative rate is equal for men and women, the expectation value for the utility is equal, and thus no group is worse off than the other in expectation.[10]

We do not claim that attributing 1 unit of disutility to all individuals who end up being wrongly denied parole is an accurate description in all cases, but it might be a reasonable and simple model for measuring disutility. Obviously, other models might lead to other fairness requirements.

In this specific scenario, equality in the false positive and false negative rates guarantees the satisfaction of (v-1-bis). Thus, if we can show that this result can be achieved through scores that are not group-calibrated, we have shown that it is possible for scores to violate between-group calibration in a way that does not fulfill (v-1-bis), thus a violation of group-calibration does not lead to prima facie wrongful discrimination, and, given our stipulation, unfairness.

Let us first of all indicate precisely what score it is that violates between-group calibration. The score in question is *R*, a score that can only take two possible values, namely 1 for "high risk of recidivism" (a positive prediction) and 0 for "low risk of recidivism" (a negative prediction). One can imagine *R* to be the result of the following process: one starts with a score *s* which is calibrated for all the relevant groups (i.e. within-group calibrated); then different thresholds are applied when computing predictions for men and women, in such a way that equality in the false positive and false negative rates can be guaranteed [12]. Or we can imagine the score $R$ to be based on a single threshold, but from a non-calibrated score $s$ that has been computed (through a suitable machine learning technique) to guarantee equality in the false positive and false negative rates of the resulting score $R$. The definition of group calibration provided in 2.2. can then of course be applied to $R$.[11] As shown by [17] and [8], if the $R$ fulfils equality in both the false-positive and the false negative rate, it is *impossible* for R to be calibrated between groups (given our stipulation of condition 4 in 4.3).

---

[9] Here we set the disutility resulting from the false-positives (wrongly releasing a future reoffending parolee) and from the true-positives (correctly denying parole to a future reoffending parolee) to be equal to zero, for both. Notice that it is the utility of the people affected by the decision, i.e., prison inmates, that is in question here, not the utility of the judge who uses the scores to take decisions. The judge, in this case, cannot be subjected to discriminatory treatment by virtue of his own decision (even if he is male, and scores for males are not within-group calibrated). If one considers the judge's utility from releasing defendants on parole, then it has to be *negative* from wrongfully releasing a parolee; but, as the argument makes clear, it is not the judge's utility that is relevant to discrimination here.

[10] Here we are assuming that we compute two expected values for the reoffending and non-reoffending populations, without taking the average, which can be given a moral justification in terms of the two groups deserving a different treatment [22]. If this assumption is dropped, the expectation value is given for each person (who may be a member of either the reoffending or non-reoffending population with a probability equal to P(Y=y)) is equal to $P[D=0|Y=1] * P(Y=1) * 1 + [D=0|Y=0] * P(Y=0) * 0$, that is to $P[D=0|Y=1] * P(Y=1)$. The expectation value is equal if and only if the value $P[D=0|Y=1] * P(Y=1)$ in the same for both groups. Now, if the false-negative rate is equal for men and women, the expectation value for the utility is *not* necessarily equal, since that equalizes $P[D=0|Y=1]$ but $P(Y=1)$ may be different in the two groups. Instead, the expectation value is equal when $P_m[D=0|Y=1] = P_f[D=0|Y=1] \cdot \frac{P_f(Y=1)}{P_m(Y=1)}$, and thus no group is worse off than the other in expectation. Since this alternative definition of parity can be achieved in the absence of calibration between groups, the general point of the paper still remains valid: weak calibration is not always necessary for fairness. We show that $P_m[D=0|Y=1] = P_f[D=0|Y=1] \cdot \frac{P_f(Y=1)}{P_m(Y=1)}$ does not logically entail calibration between groups in the Appendix.

We have therefore shown that the use of a score that violates group calibration is not necessarily unfair under our definition.

It has been argued that the violation of the false-positive and false-negative rates is *not* necessary for fairness in every case [13, 23]. We agree with this claim and our example is compatible with it. For we only show that it is *sufficient* for fairness *under the very specific definition* we give here. In this case, if the false-positive and false-negative rates are equalized, individuals of the two groups are, on average, harmed to the same extent, between-group calibration is not required.

# 7 Conclusion and Practical implications

In this paper, we have shown that a violation of group calibration may be unfair in some cases, but not unfair in others. We did so by providing a thorough moral assessment, based on a standard definition of discrimination from the philosophical literature, for two specific examples of usage of predictions in decision making.

The most important practical implication is that arguments based on examples in which fairness requires between-group calibration, or equality in the false-positive/false-negative rates, do no generalize. For it may be that group calibration is a fairness requirement in one case, but not in another. This is in line with many claims that algorithmic fairness should be defined in a way that is sensitive to context in different documents and recommendations (see, e.g. [21]). Yet, there are few other paper offering clear examples of cases in which group calibration or, alternatively, equality of false positive/false negative rate should be used, supported by a clear moral argumentation.[12]

More specifically, we show that there can *also* be cases in which a decision which equalizes the positive and negative rates (or the relevant expected utility parity condition)[13] at the expense of calibration should be regarded as fair (non-discriminatory), compatibly with what [16] also argues in some contexts, but with a very different argument. We show that equalizing false-positive and false-negative rates (or some other expected utility parity condition) may be needed to ensure that individuals of the two groups are being harmed to the same extent, on average. This means that scores 'meaning the same thing' to the decision maker[14] (the often-cited motivation for between group calibration), is not necessary for avoiding discrimination according to the account we explore.

---

[11] The coarsened score $R$ can also be interpreted as a prediction $\hat{Y}$ that the inmate will reoffend.

[12] Exceptions being [14, 22].

[13] Here we consider the condition $P_m[D = 0|Y = 1] = P_f[D = 0|Y = 1] \cdot \frac{P_f(Y=1)}{P_m(Y=1)}$ for the prisoner who may belong to either the reoffending or non-reoffending population with a probability P(Y=y), see footnote 9.

This poses the question which features of the decision context determine which fairness metric is the morally most appropriate one. It is beyond the scope of this paper to provide a rigorous methodology for this problem. However, the discussed examples suggest that recommendation systems as discussed above might be a generic setting calling for between-group calibration, especially if the recommendation is delivered as a raw score, and knowledge of the score improves the decisions of members of different groups to different degrees. In this case, the possibly discriminated persons are the direct users of the prediction model, taking decisions for themselves while relying on a specific meaning of what the score actually means. So, systematic differences in the meaning of a score between different groups are likely to become a reason for discrimination. On the other hand, in situations where a decision maker makes decision for or on other people (like in the judge case) more options are available to adapt the properties of the decision system to the moral requirements of the specific context, e.g., computing coarsened scores ("predictions") from raw scores via different threshold rules. Note that calibrated scores may nevertheless be important for the decision-maker, e.g. as a prerequisite for increasing overall welfare.[15] But increasing overall welfare is not, given our account of fairness, a requirement of fairness. As the discussed examples show, the central issue is a thorough model of what being "worse off" consists in, for a group in a given context, and to understand the mechanism how prediction and/or decision rules are related to being "worse off" in the specific context.

The limitations of our argument are the following: it deals with a very specific definition of fairness, i.e., unfairness is defined to be identical to *prima facie* morally wrongful discrimination [19]. For all we know, violating group calibration may be unfair relative to a different definition. Moreover, even if violating group calibration is not unfair in some cases, it is still possible that it may be morally bad, or impermissible, for other reasons, which are unrelated to fairness.

**ACKNOWLEDGMENT**

This work was supported by the National Research Programme "Digital Transformation" (NRP 77) of the Swiss National Science Foundation (SNSF), grant number 187473 and the European Union's Horizon 2020 research and innovation programme under the Marie Sklodowska-Curie grant agreement No 898322. The authors wish to thank Will Fleisher and three anonymous referees for helpful comments on different drafts of this paper.

---

[14] E.g., the judge, in example 2.
[15] E.g., more satisfied customers on the movie streaming platform or less offenders unnecessarily incarcerated / unsafely released.

**Appendix: proof that non-discrimination does not require calibration between groups**

We present a case where a non-calibrated (coarsened) score R is used (i.e. a score that does not fulfil $P(Y = y \mid A = a, R = r) = P(Y = y \mid R = r)$, for all values $r, a,$ and $y$), but still discrimination is avoided.

We assume that a judge starts from a calibrated score s and applies a decision rule such that the expected benefit E(U) is equal for men and women, for a given population A consisting of men and women. With the given utility function, this means that the judge makes sure that

$$P_m[D = 0, Y = 1] = P_f[D = 0, Y = 1]$$

This requirement is equivalent to

$$P_m[D = 0 | Y = 1] \cdot P_m(Y = 1) = P_f[D = 0 | Y = 1] \cdot P_f(Y = 1) \qquad (*)$$

which can also be expressed as

$$P_m[D = 0 | Y = 1] = P_f[D = 0 | Y = 1] \cdot \frac{P_f(Y=1)}{P_m(Y=1)}.$$

We show that a fulfilment of Eq. (*) does not require that

$$P_m[Y = 1 | D = 0] = P_f[Y = 1 | D = 0]. \qquad (**)$$

Proof:

We assume that the judge uses some group-dependent decision rule $D$ based on the score s, i.e. D = f(s, a) such that (*) is fulfilled. [12] have shown, for the special case of $\frac{P_f(Y=1)}{P_m(Y=1)} = 1$, that it is possible to fulfill (*) by applying a group-specific threshold rule. This can easily be generalized for other ratios $\frac{P_f(Y=1)}{P_m(Y=1)}$. Let us assume for argument's sake that Equation (**) is fulfilled.

For simplicity, we assume that the judge applies this particular decision rule of [12]. Note, however, that the proof does not depend on using this particular decision rule – it is valid also for any other decision rule fulfilling (*).

Since we assume that the score s is calibrated, we have

$$P_m[Y = 1] = \int_0^1 f_m(s) \cdot s \, ds$$

where $f_m(s)$ is the score distribution for men, i.e. $f_m(s) \, ds$ is the probability that a randomly selected man has a score in the interval [s,s+ds], and $\int_0^1 f_m(s) \, ds = 1$. This distribution $f_m(s)$ consists of two components:

$$f_m(s) = f_{m,0}(s) + f_{m,1}(s)$$

where $f_{m,0}(s)$ is the score distribution of men with Y=0, and $f_{m,1}(s)$ is the score distribution of men with Y=1.[16] These score distributions define the group of men in terms of the joint distribution of Y and s.

We have

$$P_m[Y = 1] = \int_0^1 [f_{m,0}(s) + f_{m,1}(s)] \cdot s \, ds$$
$$= \int_0^1 f_{m,0}(s) \cdot s \, ds + \int_0^1 f_{m,1}(s) \cdot s \, ds$$

If the chosen decision rule is being applied to the full population A, then a specific value of $P_m[Y = 1|D = 0]$ is obtained. Let's denote this value by $x_{A,m}$. At the same time, the application of the decision rule leads to a specific value $x_{A,f} = P_f[Y = 1|D = 0]$. Eq. (**) is fulfilled if $x_{A,m} = x_{A,f}$.

Now consider that the judge applies the same decision rule to a second population B which is identical to the first one, with the only exception that the score distribution $f_{m,0}(s)$ is different, but with the same expectation value $\int_0^1 f_{m,0}(s) \cdot s \, ds$. This means that the base rate $P_m[Y = 1]$ of the men of population B is the same as for the men of population A. As an example, Figure 2 shows two different score distributions $f_{m,0}(s)$ with the same expectation value but a different score distribution.

As nothing is changed for men with Y=1, and nothing has changed for women, all terms of Eq. (*) have the same values for population B and for population A, and if condition (*) is fulfilled for the first population (which we assumed), it is also fulfilled for the second population. So, there is no discrimination.

However, it is easy to see that $P_m[Y = 1|D = 0]$ might well be changed, as the fraction of men with Y=0 receiving a decision of D=0 is a function of the score distribution $f_{m,0}(s)$, which is different for the second population (cmp. Figure 2). The application of the judge's decision rule to population B thus results in a different value of $P_m[Y = 1|D = 0]$, say $x_{B,m}$, whereas for women we have $x_{B,f} = x_{A,f}$.

Thus, if for the original population Eq.(**) was fulfilled, this would not be the case for the second population B (and vice versa). This proves that (**) is not a necessary condition for (*) (and it makes plausible that it is very unlikely that Eq.(**) is fulfilled at all by the judge's thresholding operation).

---

[16] These are normalized such that $\int_0^1 f_{m,1}(s) \, ds = P(Y = 1)$, and $\int_0^1 f_{m,0}(s) \, ds = 1 - P(Y = 1)$.

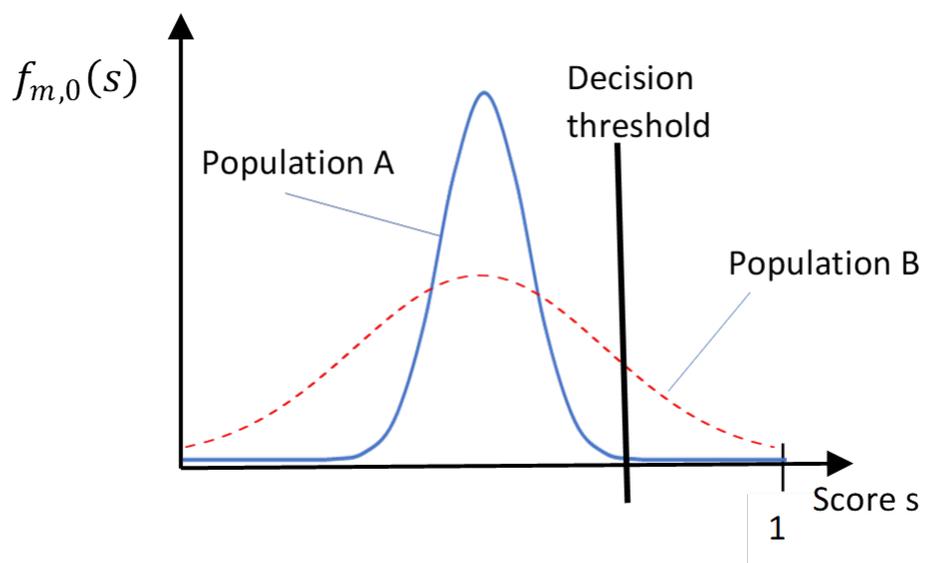

Figure 2: Score distributions $f_{m,0}(s)$ of men with Y=0, for two different populations A and B. The distributions are such that $\int_0^1 f_{m,0}(s) \cdot s \, ds$ is identical. It can be ssen that the share of men (with Y=0) receiving a decision of D=0 is different for the two populations.